# Tracking Words in Chinese Poetry of Tang and Song Dynasties with the China Biographical Database

**Chao-Lin Liu[†] and Kuo-Feng Luo[‡]**

[†]Department of East Asian Languages and Civilizations, Harvard University, USA
[†]Institute for Quantitative Social Science, Harvard University, USA
[†‡]Department of Computer Science, National Chengchi University, Taiwan
chaolinliu@fas.harvard.edu

## Abstract

Large-scale comparisons between the poetry of Tang and Song dynasties shed light on how words and expressions were used and shared among the poets. That some words were used only in the Tang poetry and some only in the Song poetry could lead to interesting research in linguistics. That the most frequent colors are different in the Tang and Song poetry provides a trace of the changing social circumstances in the dynasties. Results of the current work link to research topics of lexicography, semantics, and social transitions. We discuss our findings and present our algorithms for efficient comparisons among the poems, which are crucial for completing billion times of comparisons within acceptable time.

## 1 Introduction

Words are basic units for sentences, with which we convey ideas. Understanding the meanings carried by words, both explicitly and implicitly, is essential for correct and successful communication. The ability to "read between the lines" is important for thorough understanding. In addition to considering collocations, for Chinese, the ways a word that was commonly used and the stories that associated with certain phrases often influence an expression's connotation sensed by readers of appropriate background knowledge. For instance, "梧桐" /wu2 tong2/ [1] literally means Chinese parasol trees, but was often used in poetry about separations. Hence, "梧桐" has become a symbol of separation in literary works, similar to that "olive twigs" symbolizes peace in the Western world.

With the availability of the text files of the poetry, we can search, analyze, and compare their contents to learn about the history of word usage in the literature algorithmically. Software tools allow us to conduct research about poetry in a larger scale and from various perspectives that were practically hard for human experts to achieve before.

Studying Chinese poetry with computing technologies started at least two decades ago, so we do not mean to provide a comprehensive review of the literature. Lo and her colleagues implemented a computer assisted environment (Lo et al. 1997). Hu and Yu (2001) reported some analyses of unigrams and bigrams in Tang poems, and looked for Chinese synonyms in Tang and Song poems (Hu & Yu 2002). Lee attempted to do dependency parsing of Tang poems (Lee & Kong 2012), and explored the roles of named entities, e.g., seasons and directions, in Tang poems (Lee & Wong 2012).

We present some experiences in analyzing and comparing the contents of the Complete Tang Poem (全唐詩 /quan2 tang2 shi1/, CTP henceforth) and the Complete Song Lyrics (全宋詞 /quan2 song4 ci2/, CSL henceforth) with software tools. We choose CTP and CSL because Tang (618-907AD) and Song (960-1279AD) are arguably the most influential stages in the history of Chinese literature and because poem (詩, /shi1/) and lyrics (詞, /ci2/) are, respectively, the most representative forms of poetry in these dynasties. The influences of the poetry in these dynasties last until today. In addition, we access the China Biographical Database (Fuller 2015, CBDB henceforth) for information about the poets to enhance the overall results of our investigation. We can expand our work to cover literature of earlier and later dynasties whenever the text files and biographical data become available.

---

[1] Chinese words will be followed by their Hanyu Pinyin and tones.

We implement tools for efficient comparisons and analyses of poems and apply some freeware in our work. There are, respectively, 42,863 and 19,394 items in our CTP and CSL files. Comparing each item with others needs more than 1.9 billion comparisons. The number of comparisons will increase exponentially when we expand our study into Complete Song Poem, which has more than 185 thousand items. Hence, an efficient strategy for comparing poems is very important.

In Section 2, we provide more background information about analyzing poetry with software tools, and illustrate the benefits of considering biographical data in the analysis of literary works in Section 3. We turn our attention to algorithms for comparing the contents of poems in Section 4, and, in Section 5, we discuss some interesting findings that we noticed with the help of our tools. We briefly review some challenging issues and make concluding remarks in Section 6.

## 2 More Background Information

Software tools for textual analysis provide ample opportunities for us to study Chinese poetry from a variety of new positions. On comparing the poems of Li Bai (李白)[2] and Du Fu (杜甫), two very famous Tang poets, Jiang (2003) presented his observations from a close-reading viewpoint, and we showed the poets' differences from a distant-reading standpoint (Liu et al. 2015).

Researchers may focus their investigation on a special aspect of CTP, e.g., Pan (2015) introduced his observations about words about plants and flowers in Chinese poetry. We consider that colors portray the scenery that could be delivered by a poem; just like that audio effects drive the atmosphere in a movie. The most frequent color in CTP is white (白 /bai2/). Following this direction, we have reported some findings about poets' styles and cultural implications that are related to colors (Liu et al. 2015, Cheng et al. 2015). In addition, we found that red (紅 /hong2/) is the most frequent color in CSL (Liu 2016), and it is possible to link this observation to social and cultural circumstances of the Song dynasty. Poets, both male and female, may express themselves from female perspectives and may use females as metaphors for goals that were hard to achieve (Cheng et al. 2015, Sun 2016).

In addition to offering efficient search and comparison capabilities, software tools should facilitate the research by linking more relevant data about the poets. When studying the poems of a specific poet, a researcher should learn about the poet's life to better appreciate the meanings hidden in the poems.

We test this intuition by using the China Biographical Database (CBDB) in our work. CBDB provides information about approximately 360,000 individuals primarily from the 7th through 19th centuries in China. We demonstrate two applications of the information about the birth year, death year, and the alternative names of the poets in CBDB in the next section.

## 3 Linking Historical and Literary Analysis

### 3.1 Social Networks among Poets

Social network analysis (SNA) proves to be an effective instrument in social science studies. It is perhaps a bit surprising that researchers had attempted to study connections among poets without the assistance of modern computers (Wu 1993), although the results are not perfect.

In CTP, a poet may mention another poet's name in the title or in the content of a poem. It is not difficult to determine whom was mentioned if the complete names were used.

CBDB records the poets' alternative names, with which we can find more connections between poets. Often, the alternative names are short, containing just one or two characters, and it is not easy to pinpoint the alternative names in the contents of the poems.

We rely on some heuristics to increase the precision of our SNA analysis. For instance, we use the string of the alternative names as an evidence for the relationship between two poets only if one poet mentioned the other with the latter's full name in other poems. This design choice may hurt the recall rate, and may be adjusted if necessary.

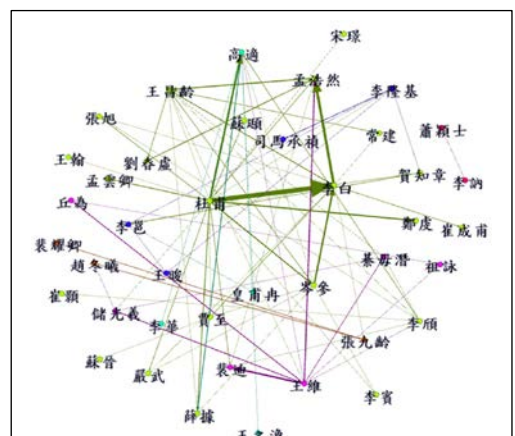

**Figure 1. Poet network for high Tang**

---

[2] The first word is the surname in Chinese names.

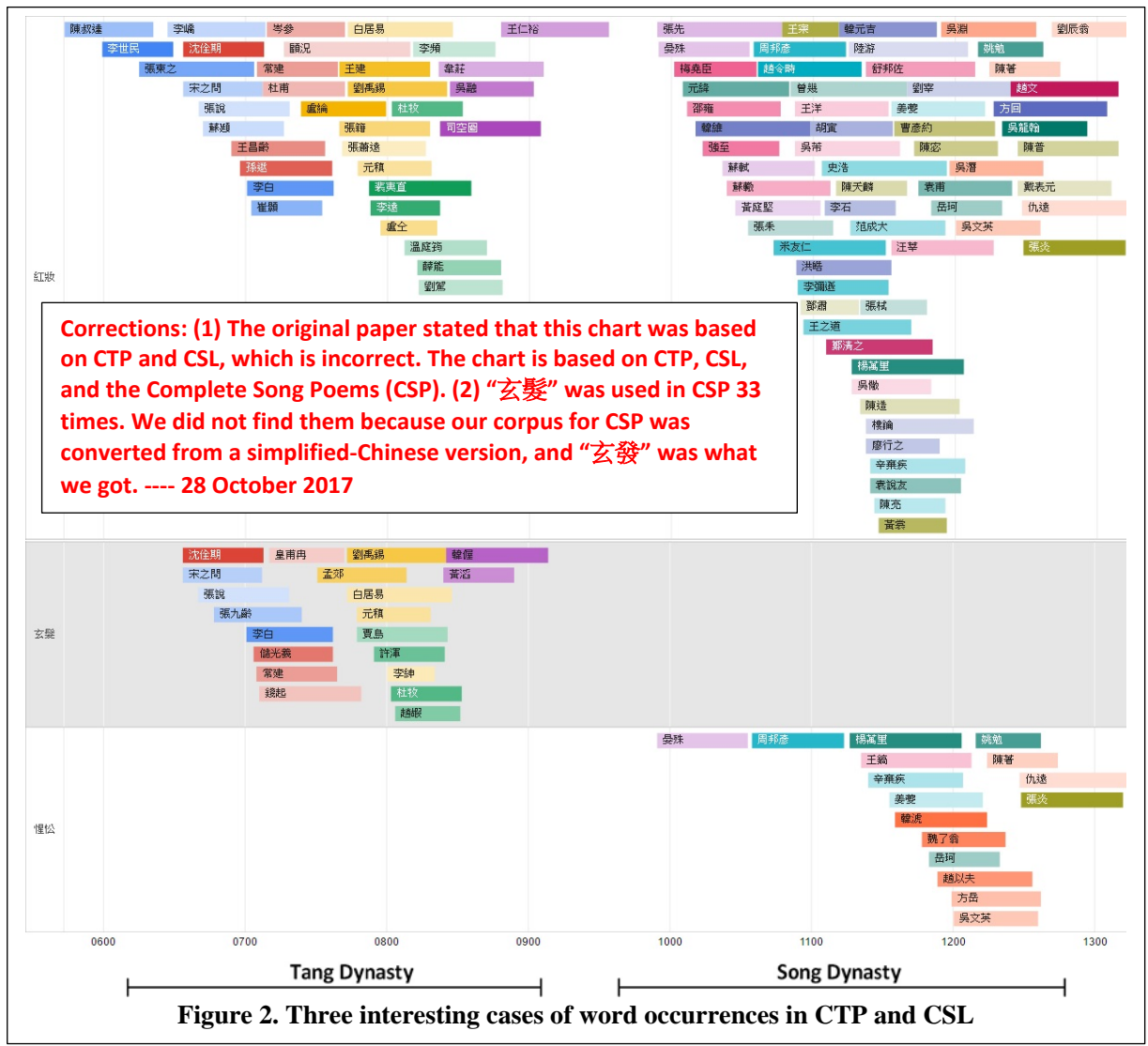

**Figure 2. Three interesting cases of word occurrences in CTP and CSL**

Figure 1 shows a social network that indicates the mentioning of poets' names for poets of the high-Tang period (713-765AD)[3]. The arrows point to the names that were mentioned in the poems (of the poets whose names are at the tails of the arrows), and thicker arrows suggest higher frequencies.

The social networks thus identified can be used for historical and literary studies. After experts verify the relationships, we can record the relationships in CBDB to enrich the contents of CBDB. One may also analyze and compare the styles and subjects of the poems of the poets who frequently mentioned each other to check, for example, whether friends had common interests in their poems.

### 3.2 A History of Word Occurrences

Compiling a comprehensive Chinese word dictionary is a huge, if not formidable, task. Luo (1986) led hundreds of scholars to achieve a contemporary version in 1986. We can enhance the lexicon with more examples from the Tang and Song poetry.

Specifically, we apply techniques of information retrieval (Manning et al. 2008) to track how words were used in Chinese literature over time. With the birth and death years of the poets that were recorded in CBDB, we can draw a chart like Figure 2 to show a history about the words[4]. The horizontal axis of Figure 2 shows the years of Tang and Song dynasties, and the widths of the

---
[3] Figure 1 was created with Gephi <https://gehpi.org>.
[4] Figure 2 was produced with the support of Google Charts <https://developers.google.com/chart/>.

> **Corrections: (1)** Because we were using CSP and CSL rather than just CSL in Figure 2, we have more poets in CSL and CSP than in CTP. The claim "Although there were … about social changes" needs to be removed in this report. **(2)** "玄髮" was used in both CTP and CSP, but was not used in CSL. ---- 28 October 2017

```
Algorithm FindCommon
  Input: 1. sets of poems S={S₁, S₂, …, Sᵢ,…,S_N}, each Sᵢ is a
            collection of poems (either CTP or CSL or others),
            i.e., Sᵢ = {P_{i,1}, P_{i,2}, …, P_{i,qi}}, where a P_{j,k} is the k-th
            poem in S_j
         2. basic filtering conditions, F
         3. output format requests, R
  Output: common parts of any two poems in S
Steps:
1  Compute an indexed list of characters, V, that are used in S
2  For any two poems, P_x and P_y, do the following.
   2.1 Look up the characters of P_x in V, and save the indexes
       for the characters in I_x. Repeat this step for P_y to create
       I_y.
   2.2 Compare the indexes in I_x and I_y to find the characters
       that appear in both P_x and P_y. Record the locations of the
       common characters in C_x and C_y, respectively.
   2.3 Emit the common words in format R, along with basic
       information about P_x and P_y, if the common words satisfy F
```

**Figure 3. Our algorithm for comparing poems**

rectangles that contain the poets' names[5] indicate the poets' life span. We do not show poets whose life spans are not known in Figure 2. The figure is divided into three parts, from top to bottom, for "紅妝" /hong2 zhuang1/, "玄髮" /xuan2 fa3/, and "惺忪" /sing1 song1/, each showing the poets who used these three words.

An interface like Figure 2 can provide useful information that a traditional lexicon may not achieve easily. First, the chart offers a distant reading of the history of the word's occurrences. Although there were more poets in CTP than in CSL, more CSL poets used "紅妝" in their works than CTP poets did, which provides hints about social changes (cf. Sun 2016). We can easily see that "玄髮" was used only in CTP and that "惺忪" might have been an invented word in the Song dynasty.

Second, we can strengthen the charts for close reading, style analysis, and other applications. Researchers can click on the poets' names to read the poems that actually used the specific words, e.g., "紅妝", for further investigation. Given the time stamps on the horizontal axis, one may study how poets used "紅妝" in a specific time period, e.g., high Tang or Southern Song periods. Maybe more interesting is that we can automatically extract the poems that used a specific word to study whether the meanings carried by the word changed over time. Moreover, for language learners, our work can serve as a source of sample poems that used selected words.

## 4 Locating Shared Words of Poems

### 4.1 Comparing Individual Poems

We design the algorithm, `FindCommon` in Figure 3, to compare large sets of poems efficiently. To simplify our illustration, we assume that there are only two items in CTP and only one item in CSL, and we refer to an individual work as a poem, temporarily ignoring whether they are Tang poems or Song lyrics.

In CTP, we have the following two poems authored by Liu Yu-Xi (劉禹錫).

$P_{11}$: 山圍故國周遭在，潮打空城寂寞回。淮水東邊舊時月，夜深還過女牆來。[6]
$P_{12}$: 朱雀橋邊野草花，烏衣巷口夕陽斜。舊時王謝堂前燕，飛入尋常百姓家。

---
[5] All Chinese characters within the boxes are poets' names, and we do not provide their Hanyu Pinyin here.
[6] We could not show the Hangyu Pinyin for the poems due to page limits. The titles of $P_{11}$, $P_{12}$, and $P_{13}$, are, respectively, "石頭城" /shi2 tou2 cheng2/, "烏衣巷" /wu1 yi1 siang4/, and "大石金陵" /da4 shih2 jin1 ling2/.

In CSL, we have the following item authored by Zhou Ban-Yan (周邦彥)

**P$_{21}$:** 佳麗地，南朝盛事誰記？山圍故國繞清江，髻鬟對起。怒濤寂寞打孤城，風檣遙度天際。
斷崖樹、猶倒倚，莫愁艇子誰係？空餘舊跡鬱蒼蒼，霧沉半壘。夜深月過女牆來，傷心東望淮水。
酒旗戲鼓甚處市？想依稀，王謝鄰裏，燕子不知何世，向尋常巷陌人家。相對如說興亡，斜陽裏。

At the first step, we scan the contents of every poem in the datasets, and record each different character in a list. The characters are indexed for efficient lookup operations, and this list serves as a basis for comparing the contents of individual poems. With the three poems, we may have a V like {"山":0, "圍":1, "故":2, …, "月":20, "夜":21, "深":22, "還":23, "過":24, "女":25, "牆":26, "來":27, …}. We chose to index at the character level so that we can find all of the shared characters in poetry.

At step 2.1, we convert a poem into a list of indexes (from V) for characters that appeared in the poem. In this illustration, I$_{11}$ will be "0, 1, 2, …, 27". P$_{21}$ is long, so I$_{21}$ will be a long list of indexes. The sentence "夜深月過女牆來" in P$_{21}$ will contribute "20, 21, 22, 24, 25, 26, 27" to I$_{21}$.

At step 2.2, we compare the lists of indexes for P$_x$ and P$_y$ to find common characters. Comparing indexes of characters is computationally more efficient than directly comparing the characters. After computing the intersection of I$_{11}$ and I$_{21}$, we can determine that "月", "夜深", "過女牆來" appeared in P$_{11}$ and P$_{21}$. Note that P$_{21}$ does not use "還", so C$_{11}$ will read like { …, "月", "夜深", "過女牆來"}. C$_{11}$ includes characters in P$_{11}$ and P$_{21}$, when we compare them. Likewise, each character in "夜深月過女牆來" of P$_{21}$ appeared in P$_{11}$, so C$_{21}$ would read like {…, "夜深月過女牆來", …}.

At step 2.3, we can select the strings that would appear in the final report. If researchers are not interested in unigrams, like "月" in this illustration. We can remove strings that are shorter than a given threshold, and this can be done via F in the input.

This example also shows us that there are at least two ways to report the common characters of two poems. In the current case, we may report different common strings, i.e., C$_{11}$ or C$_{21}$, depending on our standpoint as we just explained. This can be controled via R in the input. Notice that the choice of standpoint can have a variety of influences on the output, e.g., when we compare P$_{12}$ and P$_{21}$, C$_{12}$ and C$_{21}$ will contain "陽斜" and "斜陽", respectively.

In summary, if we compare P$_{11}$ and P$_{21}$ and report all of the common strings (including unigrams) in terms of words in P$_{21}$, we will find {"山圍故國", "寂寞打", "城", "空", "舊", "夜深月過女牆來", "東", "淮水"}. If we compare P$_{12}$ and P$_{21}$ and report all of the common strings in terms of words in P$_{21}$, we will find {"舊", "王謝", "燕", "尋常巷", "家", "斜陽" }.

We produce the following record after we compare P$_{11}$ and P$_{21}$ and report all of the common strings (including unigrams) in terms of words in P$_{21}$. In addition to the common words, we add the poet names and the IDs of the poems that are compared for each record. A record contains three fields that are separated by "|||". We put P$_{21}$ in the leftmost field because the common words, which are grouped in the rightmost field, are listed in the terms that appeared in P$_{21}$, i.e., from the standpoint of P$_{21}$.

**Zhou-Ban-Yan_P$_{21}$||| Liu-Yu-Xi_P$_{11}$|||**[山圍故國,寂寞打,城,空,舊,夜深月過女牆來,東,淮水]

**Zhou-Ban-Yan_P$_{21}$||| Liu-Yu-Xi_P$_{12}$|||**[舊,王謝,燕,家,尋常巷,斜陽]

We can offer different viewpoints for researchers to examine the words shared by the poems. Although we read "夜深月過女牆來" in P$_{21}$, this string actually came from three shorter strings in P$_{11}$. i.e., "月", "夜深", and "過女牆來". Hence, a researcher can choose to see the list of common words in the following manners, by appropriately setting R when s/he runs FindCommon.

**Zhou-Ban-Yan_P$_{21}$||| Liu-Yu-Xi_P$_{11}$|||**[山圍故國,寂寞打,城,空,舊,月,夜深,過女牆來,東,淮水]

**Liu-Yu-Xi_P$_{11}$ ||| Zhou-Ban-Yan_P$_{21}$|||**[山圍故國,打空城寂寞,淮水東,舊,月,夜深,過女牆來]

## 4.2 Selecting Interesting Candidates

We have 42,863 items in CTP and 19,394 items in CSL. An exhaustive comparison procedure that considers two viewpoints of a poem pair would conduct more than 3.8 billion comparisons in FindCommon. On one personal desktop computer with an Intel i7-4790 3.6G CPU, the Microsoft

Windows 10 64-bit Operating System, 32G RAM, and an ordinary hard disk, it took about 35 hours to complete the comparisons with our Java programs.

The computation time will increase noticeably when we include the Complete Song Poems (全宋詩 /quan2 song4 shi1/, CSP henceforth) in the comparison procedure. Like CTP and CSL, different sources of CSP may contain slightly different numbers of poems. There are more than 185 thousand items in our CSP. Comparing just one viewpoint for all items in CTP, CSL, and CSP needs more than 30 billion comparisons and will consume about 10 days with one computer.

Of course, the results of comparing any pair of poems are mutually independent, so we could and should run the comparisons in parallel on multiple machines. Nevertheless, this is a resource-consuming step, and we do not want to repeat these basic comparisons again and again.

Therefore, we organize the search for poem pairs that may have interesting common words into two stages. At the first stage, we employ `FindCommon` to compare all pairs of poems and find all common strings, including unigrams. We record the common strings of any pair of poems, except those pairs that share no or only one character, assuming that these instances are not of interest.

This, as one may expect, will produce huge output files, and, indeed, comparing just CTP and CSL will generate an output file that is larger than 300G in size. The actual size of the output file varies with `F` and `R` that we set when we run `FindCommon`.

At the second stage, a researcher will set criteria for selecting records from what we have obtained at the first stage. This will help the researcher to focus on a much smaller set of pairs of poems than those records that we obtain at the first stage. We continue to employ the previous example to illustrate the main idea.

We will obtain the following two instances when we compare $P_{11}$ and $P_{12}$ at the first stage. At the second stage, a researcher can choose to ignore both instances by asking the filter to output instances in which the list of common words has at least two bigrams. Alternatively, the researcher may choose to check instances that have at least two substrings, and, in this case, the second instance will survive.

**Liu-Yu-Xi_$P_{11}$|||Liu-Yu-Xi_$P_{12}$|||[邊舊時]**
**Liu-Yu-Xi_$P_{12}$|||Liu-Yu-Xi_$P_{11}$|||[舊時, 邊]**

## 5  Shared Texts among Poetry of Tang and Song Dynasties

We discuss some interesting instances in which terms, sentences, or imageries were shared among Tang and Song poetry in this section (cf. Wang 2003). Although our findings can lead to several types of further investigations, we present samples that roughly fall into two categories. The shared words can nurture certain similar or related imagery in poems, and the shared words and expressions may suggest some authorship or version issues of the poetry.

The running example that we elaborated in previous section is a famous example of using several terms from multiple sources in a new poem (cf. Chen & Wang 2001). In a more complete account, Zhou Ban-Yan also used a poem of Xie Tiao (謝朓) and a Yuefu poem (樂府詩)[7] in $P_{21}$. We did not discuss these additional poems partially because they are not part of CTP or CSL.

We summarize the results of the comparisons in Section 4 in the following manner. We mark shared characters with tiny ripples under them. The shared characters are colored in green for items in CTP and in blue for items in CSL. The original poems are shown along with poets' names on the left.

**Liu Yu-Xi**: 山圍故國周遭在，潮打空城寂寞回。淮水東邊舊時月，夜深還過女牆來。(CTP)
**Liu Yu-Xi**: 朱雀橋邊野草花，烏衣巷口夕陽斜。舊時王謝堂前燕，飛入尋常百姓家。(CTP)
**Zhou Ban-Yan**: 佳麗地，南朝盛事誰記？山圍故國繞清江，髻鬟對起。怒濤寂寞打孤城，風檣遙度天際。斷崖樹、猶倒倚，莫愁艇子誰係？空餘舊跡鬱蒼蒼，霧沉半壘。夜深月過女牆來，傷心東望淮水。酒旗戲鼓甚處市？想依稀，王謝鄰裏，燕子不知何世，向尋常巷陌人家。相對如說興亡，斜陽裏。(CSL)

Sometimes, poets would directly reuse the same sentences that had been used in other poems. In CSL, He Zhu (賀鑄) reused two sentences in a poem of Du Mu (杜牧) in CTP.

---

[7] Xie Tiao: 江南佳麗地，金陵帝王州。逶迤帶綠水，迢遞起朱樓。飛甍夾馳道，垂楊蔭御溝。凝笳翼高蓋，疊鼓送華輈。獻納雲臺表，功名良可收。Yuefu: 莫愁在何處，莫愁石城西；艇子打兩槳，催送莫愁來。

**Du Mu:** 清時有味是無能，閑愛孤雲靜愛僧。欲把一麾江海去，樂游原上望昭陵。(CTP)

**He Zhu:** 閑愛孤雲靜愛僧，得良朋。清時有味是無能，矯聾丞。況復早年豪縱過，病嬰仍。如今痴鈍似寒蠅，醉懵騰。(CSL)

In another example, He Zhu reorganized a few terms of Li Shang-Yin (李商隱) in his own poem.

**Li Shang-Yin:** 為有雲屏無限嬌，鳳城寒盡怕春宵。無端嫁得金龜婿，辜負香衾事早朝。(CTP)

**He Zhu:** 章台遊冶金龜婿。歸來猶帶醺醺醉。花漏怯春宵。雲屏無限嬌。絳紗燈影背。玉枕釵聲碎。不待宿酲銷。馬嘶催早朝。(CSL)

The follow example shows that He Zhu shared words with three poets: Zhang Ji (張籍), Xu Hun (許渾), and Cui Tu (崔塗) in one poem.

**Zhang Ji:** 青山歷歷水悠悠，今日相逢明日秋。系馬城邊楊柳樹，為君沽酒暫淹留。(CTP)

**Xu Hun:** 紅花半落燕於飛，同客長安今獨歸。一紙鄉書報兄弟，還家羞著別時衣。(CTP)

**Cui Tu:** 海棠花底三年客，不見海棠花盛開。卻向江南看圖畫，始慚虛到蜀城來。(CTP)

**He Zhu:** 排辦張燈春事早。十二都門。物色宜新曉。金犢車輕玉驄小。拂頭楊柳穿馳道。薦羹鱸鱠非吾好。去國謳吟，半落江南調。滿眼青山恨西照。長安不見令人老。(CSL)

`FindCommon` would also discover Xin Qi-Ji (辛棄疾) and Wun Bing (文丙) shared some words in their poems.

**Wun Bing:** 可憐同百草，況負雪霜姿。歌舞地不尚，歲寒人自移。階除添冷淡，毫末入思惟。盡道生雲洞，誰知路嶮巇。(CTP)

**Xin Qi-Ji:** 暗香橫路雪垂垂。晚風吹。曉風吹。花意爭春，先出歲寒枝。畢竟一年春事了，緣太早，卻成遲。未應全是雪霜姿。欲開時。未開時。粉面朱唇，一半點胭脂。醉裡謗花花莫恨，渾冷淡，有誰知。(CSL)

It is certainly possible for us to compare poems in CTP. We could find the following two items that were listed under the names of different authors, with very different titles, and in two different volumes. The names of the poets are Lu Lun (盧綸) and Lu Shang-Shu (盧尚書). Despite these differences, the poems are extremely similar, and differ only in one character, which we show in red and mark with an under ripple.

**Lu Lun**: 夕照臨窗起暗塵，青松繞殿不知春。君看白髮誦經者，半是宮中歌舞人。(CTP)

**Lu Shang-Shu**: 夕照紗窗起暗塵，青松繞殿不知春。君看白首誦經者，半是宮中歌舞人。(CTP)

Is it possible that Lu Shang-Shu is Lu Lun and that Lu Lun revised his own work? According to the biographical information of Lu Lun, he once served as the head of "戶部" /hu4 bu4/, which was called Shang-Shu ("尚書"). From this perspective, it is possible that this Lu Shang-Shu is Lu Lun, but we will not elaborate on this issue here.

We show two more pairs of poems in CTP whose authors might be the same below. In the following pair, the poems are similar, but their titles ("別佳人" /bie2 jia1 ren2/ vs. "別妻" /bie2 ci1/) are related yet different. The names of their authors are different but could be pronounced similarly.

**Cui Ying** (崔膺)**:** 壟上流泉壟下分，斷腸嗚咽不堪聞。嫦娥一入月中去，巫峽千秋空白雲。(CTP)

**Cui Ya** (崔涯)**:** 隴上泉流隴下分，斷腸嗚咽不堪聞。嫦娥一入月中去，巫峽千秋空白雲。(CTP)

The following poems of Lu and Luo differ in just one character. They have the same title and the pronunciations of the names of their authors are very similar.

**Lu Yin** (盧殷): 累年無的信，每夜夢邊城。袖掩千行淚，書封一尺情。(CTP)

**Luo Yin** (羅隱): 累年無的信，每夜望邊城。袖掩千行淚，書封一尺金。(CTP)

The following two poems in CTP also differ in only one character. The poets are Zhang Ba-Yuan (張八元) and Zhu Fang (朱放), two really different persons, and the titles of the poems are the same. We checked the pages of a hard copy of the CTP, and verified the different characters, i.e., "夫" /fu1/ and "天" /tian1/. Hence, we have identified another type of authorship problem.

**Zhang Ba-Yuan:** 昨辭夫子棹歸舟，家在桐廬憶舊丘。三月暖時花競發，兩溪分處水爭流。近聞江老傳鄉語，遙見家山減旅愁。或在醉中逢夜雪，懷賢應向剡川遊。(CTP)

**Zhu Fang:** 昨辭天子棹歸舟，家在桐廬憶舊丘。三月暖時花競發，兩溪分處水爭流。近聞江老傳鄉語，遙見家山減旅愁。或在醉中逢夜雪，懷賢應向剡川遊。(CTP)

The following poems in CTP show yet another type of challenge. The poets are Dai Shu-Lun (戴叔倫), Qing Jiang (清江), and Ke Zhi (可止). Dai was the eldest, and Ke was born at least 50 years after Qing deceased. The titles and contents of Qing's and Ke's poems were exactly the same. The title of Dai's poem is different. Moreover, both Qing's and Ke's poems differ from Dai's in just one character.

**Dai Shu-Lun**: 空門寂寂澹吾身，溪雨微微洗客塵。臥向白雲晴未盡，任他黃鳥醉芳春。(CTP)
**Qing Jiang**: 空門寂寂淡吾身，溪雨微微洗客塵。臥向白雲情未盡，任他黃鳥醉芳春。(CTP)
**Ke Zhi**: 空門寂寂淡吾身，溪雨微微洗客塵。臥向白雲情未盡，任他黃鳥醉芳春。(CTP)

## 6  Discussions and Concluding Remarks

Implied meanings of words and collocations could vary from poet to poet and from dynasty to dynasty. Connotation and imagery associated with words in poetry still show their influences in modern Chinese. We design `FindCommon` to identify and show the poetry that contained the shared words and collocations, and discuss possible applications of such findings. In addition to the Complete Song Poem, we certainly can and should extend the current work to include earlier Chinese poetry, i.e., Shijing (詩經), Verses of Chu (楚辭), and Hangfu (漢賦), and later ones, e.g., Complete Qing Poem (Zhu 1994) to accomplish a more complete history of words and collocations in Chinese poetry.

Results of the current work can be improved if we can achieve high-quality word segmentation in poetry. The quality of the corpora based on which we conduct the comparisons directly affects researchers' observations. As long as we can obtain more reliable and authoritative corpora, we can rerun the analysis and offer better services to humanities researchers.

**Responses to Reviewers' Comments**

1. Tang and Song are two major dynasties in China's history. The spans of Tang and Song are, respectively, 618-907AD and 960-1279AD, which are now marked in Figure 2.

2. We implemented the algorithm that is listed in Figure 3. It is currently designed to compare Chinese poetry, but we could revise it to handle poetry of other languages.

3. We provided the Hanyu Pinyin for the Chinese words in this paper, except those appeared in Figure 2 and Section 4. The majority of Chinese words in Figure 2 are poets' names. In this Section 4, we discussed words used in poems and listed the poems that we compared. Words and sentences in poems generally convey imagery that is beyond their literal meanings, and it would take many words and require significant background information to appropriately translate the words and poems. Translating poems is a huge task and requires a lifetime dedication, see for example (Owen 2016). Hence, we chose to list just the words, which might be acceptable, because we focused on literal comparisons between poems in this paper.

4. The algorithm `FindCommon` listed in Figure 3 can produce all co-occurrences, though the actual output depends on the settings of `F` ad `R`. The types of co-occurrences that can be identified and presented to a researcher for further inspection will then depend on the filtering procedure that we outlined in Section 4.2. It is the researcher's judgement as to what types of co-occurrences that will be examined in the research.

**Acknowledgements**

We thank Lik Hang Tsui, Hongsu Wang, and Shuhua Zhang of the CBDB project for the informative conversations with the first author. We also thank the anonymous reviewers for their helpful comments and suggestions. This research was supported in part by the grant MOST-104-2221-E-004-005-MY3 from the Ministry of Science and Technology of Taiwan, the grant USA-HAR-105-V02 of the Top University Strategic Alliance, and the Senior Fulbright Research Grants 2016-2017.

## References

You-Bing Chen (陈友冰) and De-Shou Wang (王德寿). 2001. *Selected Appreciation of Song Lyrics*: *Northern Song* (宋詞清賞、北宋篇), 138–139, Chung Cheng Bookstore (中正書局). (in Chinese)

Wen-Huei Cheng (鄭文惠), Chao-Lin Liu, Wen-Yun Chiu, and Chu-Ting Hsu. 2015. Phenomenology of emotion politics of color: Digital humanities research on the lyrical genealogy of 'White' in the poetry of middle Tang dynasty, *Proc. of the 6th Int'l Conf. on Digital Humanities and Digital Archives*, 481–522.

Michael A. Fuller. 2015. *The China Biographical Database User's Guide*, Harvard University. <http://projects.iq.harvard.edu/cbdb/home>

Junfeng Hu (胡俊峰) and Shiwen Yu (俞士汶). 2001. The computer aided research work of Chinese ancient poems, *ACTA Scientiarum Naturalium Universitatis Pekinensis*, 37(5):725–733. (in Chinese)

Junfeng Hu and Shiwen Yu. 2002. Word meaning similarity analysis in Chinese ancient poetry and its applications, *J. of Chinese Information Processing* (中文信息学报), 16(4):39–44. (in Chinese)

Shao-Yu Jiang (蔣紹愚). 2003. "Moon" and "Wind" in Li Bai's and Du Fu's poems – Using computers for studying classical poems, *Proc. of the 1st Int'l Conf. on Literature and Information Technologies*. (in Chinese)

John Lee and Yin Hei Kong. 2012. A dependency treebank of classical Chinese poems, *Proc. of the 2012 Conf. of the North American Chapter of the Association for Computational Linguistics: Human Language Technologies*, 191–199.

John Lee and Tak-sum Wong. 2012. Glimpses of ancient China from classical Chinese poems, *Proc. of the 24th Int'l Conf. on Computational Linguistics*, posters, 621–632.

Chao-Lin Liu. 2016. Quantitative analyses of Chinese poetry of Tang and Song dynasties: Using changing colors and innovative terms as examples, *Proc. of the 2016 Int'l Conf. on Digital Humanities*, 260–262.

Chao-Lin Liu, Hongsu Wang, Chu-Ting Hsu, Wen-Huei Cheng, and Wei-Yun Chiu. 2015. Color aesthetics and social networks in complete Tang poems: Explorations and discoveries, *Proc. of the 29th Pacific Asia Conf. on Language, Information and Computation*, 132–141.

Fengju Lo (羅鳳珠), Yuanping Li, and Weizheng Cao. 1997. A realization of computer aided support environment for studying classical Chinese poetry, *J. of Chinese Information Processing*, 1:27–36. (in Chinese)

Zhufeng Luo (罗竹风, chief editor). 1986. *Comprehensive Chinese Word Dictionary* (汉语大辞典), Shanghai Cishu Publisher (上海辞书出版社). (in Chinese) <http://hd.cnki.net/kxhd/>

Christopher D. Manning, Prabhakar Raghavan, and Hinrich Schütze. 2008. *Introduction to Information Retrieval*, Cambridge University Press.

Stephen Owen. 2015. *The Poetry of Du Fu*, The Series of Library of Chinese Humanities, De Gruyter. open access: <https://www.degruyter.com/view/product/246946>

Fuh-Jiunn Pan (潘富俊). 2015. *Plants in Classic Chinese Literature* (草木緣情:中國古典文學中的植物世界), The Commercial Press (商務印書館). (in Chinese)

Yan-Hong Sun (孙艳红). 2016. Expressive styles of the Tang and Song Lyrics (唐宋词本体特征的表现形式), *Chinese Social Sciences Today* (中国社会科学网-中国社会科学报), 5 July 2016. (in Chinese)

Wei-Yung Wang (王偉勇). 2003. *A Study on the Comparisons of Tang and Song Poetry* (宋詞與唐詩之對應研究), Wen-Shi-Zhe Publisher (文史哲出版社). (in Chinese)

Ru-Yu Wu (吳汝煜). 1993. *Index for Communication Poems of Tang and Wu-Dai* (唐五代人交往詩索引), Shanghai Guji Publisher (上海古籍出版社). (in Chinese)

Ze-Jie Zhu (朱则杰). 1994. Establishing the editorial board for the Complete Qing Poem (全清诗边篹筹备委员会成立), Studies in Qing History (清史研究), 0(3):96.